\documentclass[journal]{IEEEtran}
\usepackage{cite}

\usepackage{chngpage}
\usepackage{amsmath,amssymb,amsthm}
\usepackage{algorithm,algorithmic,amssymb,amstext,enumerate,amsfonts,url,multirow,array,stfloats,flushend,multirow}
\usepackage{caption}
\usepackage{booktabs}
\usepackage{threeparttable}

\usepackage[colorlinks,linkcolor=blue,citecolor=blue,urlcolor=blue]{hyperref}

\usepackage[T1]{fontenc}

\renewcommand\arraystretch{1.3}

\def\x{{\mathbf x}}

\def\K{{\mathbf K}}

\DeclareMathAlphabet{\mathcal}{OMS}{cmsy}{m}{n}

\newcolumntype{I}{!{\vrule width 3pt}}
\newlength\savedwidth

\newlength\savewidth
\newcommand\shline{\noalign{\global\savewidth\arrayrulewidth
                            \global\arrayrulewidth 1pt}%
                   \hline
                   \noalign{\global\arrayrulewidth\savewidth}}

\ifCLASSINFOpdf
  \usepackage[pdftex]{graphicx}
\else
\fi
\usepackage{amsmath,color}
\usepackage{algorithm}
\usepackage{algorithmic}

\usepackage{array}

\usepackage{bigstrut,bigdelim,multirow}

\begin{document}
%
\title{Learning the kernel matrix by resampling}
%
%
%

\author{Xiao-Lei~Zhang
        \thanks{
Xiao-Lei Zhang is with the Center for Intelligent Acoustics and Immersive Communications, School of Marine Science and Technology, Northwestern Polytechnical University, Xi'an, China (e-mail: xiaolei.zhang9@gmail.com \textit{or} xiaolei.zhang@nwpu.edu.cn). }
}

%
%

\markboth{Journal of \LaTeX\ Class Files,~Vol.~14, No.~8, August~2017}%
{Shell \MakeLowercase{\textit{et al.}}: Bare Demo of IEEEtran.cls for IEEE Journals}
%



\maketitle

\begin{abstract}
In this abstract paper, we introduce a new kernel learning method by a nonparametric density estimator. The estimator consists of a group of k-centroids clusterings. Each clustering randomly selects data points with randomly selected features as its centroids, and learns a one-hot encoder by one-nearest-neighbor optimization. The estimator generates a sparse representation for each data point. Then, we construct a nonlinear kernel matrix from the sparse representation of data. One major advantage of the proposed kernel method is that it is relatively insensitive to its free parameters, and therefore, it can produce reasonable results without parameter tuning. Another advantage is that it is simple. We conjecture that the proposed method can find its applications in many learning tasks or methods where sparse representation or kernel matrix is explored. In this preliminary study, we have applied the kernel matrix to spectral clustering. Our experimental results demonstrate that the kernel generated by the proposed method outperforms the well-tuned Gaussian RBF kernel. This abstract paper is used to protect the idea, full versions will be updated later.
\end{abstract}


%
\IEEEpeerreviewmaketitle

\section{Introduction}

Learning data representations is an important issue for machine learning. One type of data representations are produced by hand-crafted features, such as various kinds of filters and kernels. Another type are produced by density estimators, such as k-means, Gaussian mixture models, kernel learning, etc. This paper focuses on kernel methods. Common predefined kernel functions include the linear kernel, polynomial kernels, Gaussian RBF kernels, etc. Some kernel matrices are produced by density estimators, such as SDP or metric learning.

In this paper, we introduce a new kernel learning method, which generates a kernel matrix from a simple nonparametric density estimator. The estimator consists of a group of k-centroids clusterings. Each clustering randomly selects data points with randomly selected features as its centroids, and learns a one-hot encoder by one-nearest-neighbor optimization. The estimator generates a sparse representation for each data point. Then, the nonlinear kernel matrix is constructed from the sparse representation of data. One major advantage of the proposed kernel method is that it can produce reasonable result without parameter tuning, compared to traditional nonlinear kernels, such as Gaussian RBF. Another advantage is that it is simple. We believe that the proposed method can find its applications in many learning tasks or methods where sparse representation or kernel matrix is explored, such as unsupervised learning, semisupervised learning, supervised learning. In this initial study, we have applied the kernel matrix to spectral clustering. Our experimental results demonstrate that the kernel generated by the proposed method outperforms the well-tuned Gaussian RBF kernel.

\section{Method}
Given a $d$-dimensional input data set $\mathcal{X} = \left\{\mathbf{x}_1,\ldots,\mathbf{x}_n \right\}$, the method trains $V$ ($V\gg 1$) $k$-centroids clusterings.
 For training each layer  either from the lower layer or from the original data space, we simply need to focus on training each $k$-centroids clustering, which consists of the following three steps:

 \begin{itemize}
 \itemsep=0.0pt
   \item \textbf{Random feature selection.} The first step randomly selects $\hat{d}$ dimensions of $\mathcal{X}$ ($\hat{d}\le d$) to form a subset of $\mathcal{X}$, denoted as $\hat{\mathcal{X}} = \left\{\hat{\mathbf{x}}_1,\ldots,\hat{\mathbf{x}}_n \right\}$.
    \item \textbf{Random sampling.} The second step randomly selects $k = \lfloor  \delta n \rfloor$ data points from $\hat{{\mathcal{X}}}$ as the $k$ centroids of the clustering, denoted as $\{\mathbf{w}_{1},\ldots,\mathbf{w}_{k} \}$, where $\delta \in (0,1)$ is a free parameter.
   \item \textbf{Sparse representation learning.} The third step assigns each input data point $\hat{\mathbf{x}}$ to one of the $k$ clusters and outputs a $k$-dimensional indicator vector $\mathbf{h} = [h_1,\ldots,h_k]^T$, where operator $^T$ denotes the transpose of vector. For example, if $\hat{\mathbf{x}}$ is assigned to the second cluster, then $\mathbf{h} = [0,1,0,\ldots,0]^T$. The assignment is calculated according to the similarities between $\hat{\mathbf{x}}$ and the $k$ centroids,
 in terms of some
 predefined similarity metric at the original data space, such as the squared Euclidean distance $ \arg\min_{i=1}^{k}\|\mathbf{w}_i-\hat{\mathbf{x}}\|^2$, or in terms of $\arg\max_{i=1}^{k}\mathbf{w}_i^T\hat{\mathbf{x}}$ at all other hidden layers.
 \item \textbf{Kernel matrix construction.} The last step constructs the similarity matrix $\K$ by $K_{i,j}= \x^T_i\x_j,\forall i=1,\ldots,n \mbox{ and } \forall j=1,\ldots, n$. It is obvious that $\K$ is a kernel matrix.
 \end{itemize}

\section{Applications}

We apply the kernel matrix to spectral clustering \cite{ng2001spectral}. To prevent the local minima of $k$-means clustering in spectral clustering, we run the $k$-means clustering multiple times (in this paper, 50 times), and pick the clustering result that corresponds to the lowest objective value among the candidate objective values as the final clustering result of the spectral clustering.

\section{Experiments}

The data sets for evaluation were summarized in Table \ref{table:data_set_info}. For each data set, we ran the spectral clustering 10 times and reported the average performance. The clustering results were evaluated by NMI and clustering accuracy (ACC).

We compared the proposed method with the RBF kernel based spectral clustering. For the proposed method, parameter $\delta$ was searched in grid through $[0.1:0.1:0.9]$ where the symbol $[A:b:C]$ denotes a set of numbers starting from $A$ and ending by $C$ with an interval of $b$; parameter $a$ was fixed to $0.5$; parameter $V=400$.\footnote{In our previous study, the method is insensitive to the selection of parameters $a$ and $V$ as if $a>0.3$ and $V>100$.} For the Gaussian RBF kernel, the kernel width $\sigma$ was set from $2^{[-4:1:4]}A$ where $A$ is the average of the pairwase Euclidean distances between data points.

\begin{figure*}
\resizebox{18cm}{!}{\includegraphics*{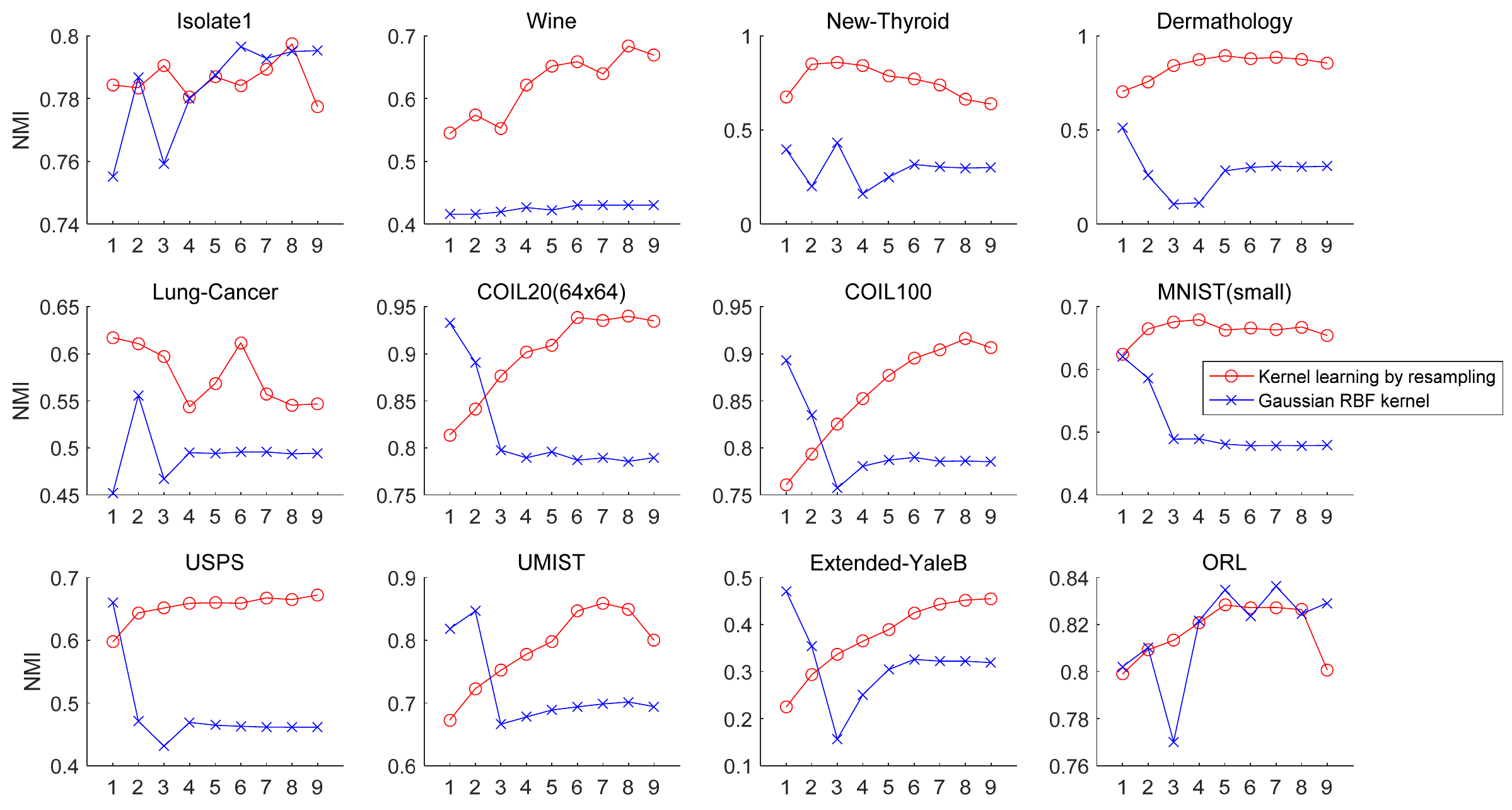}}
\caption{NMI Comparison of the proposed method and the Gaussian RBF kernel. Note that X-axis consists of the indices of the free parameters of either the proposed method or the Gaussian kernel. For the proposed method, the numbers $1,2,\ldots,9$ indexes the values $0.1, 0.2, \ldots, 0.9$ of the free parameter $\delta$ respectively. For the Gaussian RBF kernel, the numbers $1,2,\ldots,9$ indexes the values $2^{-4}A, 2^{-3}A, \ldots, 2^{4}A$ of the free parameter $\sigma$ respectively. }
 \label{fig:1}
\end{figure*}

\begin{figure*}
\resizebox{18cm}{!}{\includegraphics*{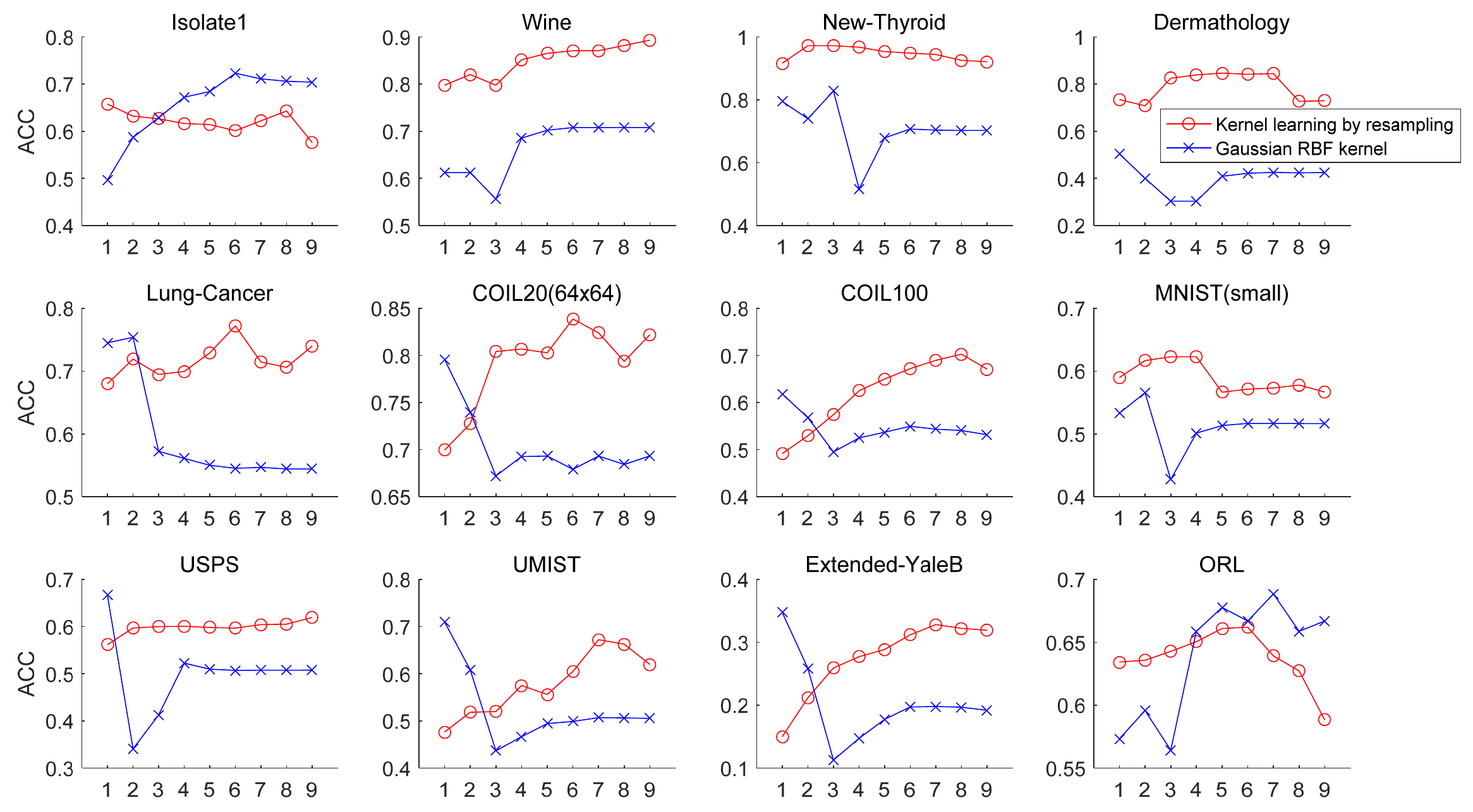}}
\caption{ACC Comparison of the proposed method and the Gaussian RBF kernel.}
 \label{fig:2}
\end{figure*}

Experimental comparison results with the Gaussian RBF kernels are shown in Figs. \ref{fig:1} and \ref{fig:2}. From the figures, we can see that, the proposed method generally reaches the optimal performance when $\delta\in[0.6, 0.8]$. This empirical conclusion is invalid apparently only on ``New-Thyroid'' and ``Lung-Cancer''. On the other side, the Gaussian RBF kernel reaches the optimal performance when $\sigma\in [2^{-4}A, 2^{-3}A]$\footnote{We failed to do eigenvalue decomposition when $\sigma <2^{-4}A$.}, and this empirical conclusion is invalid apparently on ``Isolate1'' and ``ORL''.

For unsupervised learning and clustering, because we usually do not have prior knowledge on selecting optimal free parameters, we have to fix the free parameters without manual tuning. Here, we selected $\delta = 0.7$ for the proposed method, and $\sigma = 2^{-4}A$ for the Gaussian RBF kernel. The comparison results were summarized in Tables \ref{table:clustering results_NMI} and \ref{table:clustering results_ACC} where the best performance with the optimal free parameters was also reported. The statistical difference was evaluated by the two-tailed $t$-test where the $p$-value was set to $0.05$. Comparison results show that, given the free parameters that are no matter fixed or well-tuned, the proposed method outperforms the Gaussian RBF kernel in most of the data sets, particularly in terms of NMI.

We also compared the proposed method with $\delta$ fixed to 0.7 with the well-tuned Gassian RBF kernel. From the comparison results in Tables \ref{table:NMI_small} and \ref{table:ACC_small}, we found that the the proposed method even without parameter tuning is comparable the well-tuned Gassian RBF kernel.

   \begin{table}[t]
\caption{\label{table:data_set_info}{Description of data sets.}}
\renewcommand{\arraystretch}{1.5}
\centerline{\scalebox{0.7}{
\begin{threeparttable}
\begin{tabular}{llllll}
\hline
ID &{Name} & {\# data points} & \# dimensions & \# classes & Attribute\\
 \shline
1& Isolet1 & 1560 & 617 & 26 & Speech data\\
2& Wine & 178 & 13 & 3 & Chemical data\\
3& New-Thyroid & 215 & 5 & 3 & Biomedical data\\
4& Dermathology & 366 & 34 & 6 & Biomedical data\\
5& Lung-Cancer & 203 & 12600 & 5 & Biomedical data\\
6& COIL20(64x64) & 1440  & 4096 & 20 & Images\\
7& COIL100 & 7200   & 1024  & 100 & Images\\
8& MNIST(small) & 5000  & 768 & 10 & Images (handwritten digits)\\
9& USPS & 11000  & 1024 & 10 & Images (handwritten digits)\\
10& UMIST & 575 & 1024 & 20 & Images (faces)\\
11& Extended-YaleB & 2414 & 1024 &38 & Images (faces)\\
12& ORL & 400  & 1024 &40 & Images (faces)\\
\hline
\end{tabular}
\end{threeparttable}
}}
\end{table}

   \begin{table*}[t]
\caption{\label{table:clustering results_NMI}{NMI on 12 data sets. The number after the symbol $\pm$ is the standard deviation.}}
\renewcommand{\arraystretch}{1.5}
\centerline{\scalebox{0.7}{
\begin{tabular}{llll|ll|ll}
\hline
& & {$k$-means} & $k$-means+PCA & Spectral+Gaussian\_kernel$^{no\_tuning}$& Proposed$^{no\_tuning}$ & Spectral+Gaussian\_kernel{$^{optimal}$}& Proposed{$^{optimal}$}\\
 \shline
1& Isolet1 & {77.21\%$\pm$0.92\%}& 56.74\%$\pm$0.75\%&  75.51\%$\pm$0.58\% &\textbf{78.93\%$\pm$0.44\%} & \textbf{79.66\%$\pm$0.58\%} & \textbf{79.74\%$\pm$0.39\%}\\
2& Wine&  42.88\%$\pm$0.00\%& 40.92\%$\pm$0.00\%&    41.58\%$\pm$0.00\%    &\textbf{63.94\%$\pm$0.00\%}& 43.02\%$\pm$0.00\% & \textbf{68.36\%$\pm$0.00\%}\\
3& New-Thyroid&  49.46\%$\pm$0.00\%& 49.46\%$\pm$ 0.00\%&  39.63\%$\pm$0.00\%     &\textbf{74.08\%$\pm$0.00\%}& 43.20\%$\pm$0.00\% & \textbf{85.91\%$\pm$0.00\%}\\
4& Dermathology & 9.11\%$\pm$0.11\%& 59.50\%$\pm$0.10\%&   51.04\%$\pm$0.15\%     &\textbf{88.61\%$\pm$0.00\%}& 51.04\%$\pm$0.15\% & \textbf{89.42\%$\pm$0.00\%}\\
5& Lung-Cancer & {48.59\%$\pm$1.07\%} & {49.17\%$\pm$ 0.96\%}&   45.18\%$\pm$4.60\%      &\textbf{55.70\%$\pm$0.00\%}& 55.58\%$\pm$4.60\% & \textbf{61.70\%$\pm$0.00\%}\\
6& COIL20(64x64) & 78.03\%$\pm$1.14\%& 79.00\%$\pm$1.35\%&  \textbf{93.29\%$\pm$0.81\%}       &\textbf{93.54\%$\pm$1.26\%}& \textbf{93.29\%$\pm$0.81\%} & \textbf{93.97\%$\pm$0.83\%}\\
7& COIL100 & 76.98\%$\pm$0.27\%& 69.64\%$\pm$0.45\%&   {89.28\%$\pm$0.59\%}       &\textbf{90.45\%$\pm$0.25\%}& 89.28\%$\pm$0.59\% & \textbf{91.58\%$\pm$0.52\%}\\
8& MNIST(small) & 49.69\%$\pm$0.14\%& 27.86\%$\pm$0.08\%&    62.06\%$\pm$0.04\%     &\textbf{66.30\%$\pm$0.01\%}& 62.06\%$\pm$0.04\% & \textbf{67.94\%$\pm$0.01\%}\\
9& USPS & 43.63\%$\pm$1.78\%& 43.00\%$\pm$0.05\%&      66.05\%$\pm$2.38\%      &\textbf{66.75\%$\pm$0.84\%}& 66.05\%$\pm$0.18\%  &\textbf{67.21\%$\pm$0.28\%}\\
10& UMIST & 65.36\%$\pm$1.21\%& 66.25\%$\pm$1.10\%&    {81.83\%$\pm$0.58\%}     &\textbf{85.88\%$\pm$1.65\%}& \textbf{84.66\%$\pm$2.38\%} &\textbf{85.88\%$\pm$1.65\%}\\
11& Extended-YaleB & 12.71\%$\pm$0.63\%& 16.54\%$\pm$0.56\%& \textbf{47.11\%$\pm$0.58\%}   &{44.30\%$\pm$0.72\%}& \textbf{47.11\%$\pm$0.58\%} &{45.49\%$\pm$0.58\%}\\
12& ORL & 75.55\%$\pm$1.36\%& 75.81\%$\pm$1.17\%&    {80.21\%$\pm$1.13\%}        &\textbf{82.72\%$\pm$0.97\%}& \textbf{83.62\%$\pm$1.13\%} &\textbf{82.83\%$\pm$0.61\%}\\
\hline
summary&&&&&win:10; tied:1; lose:1&&win:7; tied:4; lose:1\\
\hline
\end{tabular}
}}
\end{table*}

   \begin{table*}[t]
\caption{\label{table:clustering results_ACC}{Clustering accuracy on 16 data sets.}}
\renewcommand{\arraystretch}{1.5}
\centerline{\scalebox{0.7}{
\begin{threeparttable}
\begin{tabular}{llll|ll|ll}
\hline
& & {$k$-means} & $k$-means+PCA & Spectral+Gaussian\_kernel$^{no\_tuning}$& Proposed$^{no\_tuning}$ & Spectral+Gaussian\_kernel{$^{optimal}$}& Proposed{$^{optimal}$}\\
 \shline
1& Isolet1 & {61.47\%$\pm$1.93\%}& 38.62\%$\pm$0.99\%& 49.63\%$\pm$1.88    &\textbf{62.22\%$\pm$1.05\%} & \textbf{72.29\%$\pm$1.92\%} & 65.77\%$\pm$1.63\% \\
2& Wine& 70.22\%$\pm$0.00\%& 78.09\%$\pm$0.00\%&  61.24\%$\pm$0.00     &\textbf{87.08\%$\pm$0.00\%} & 70.79\%$\pm$0.00\% & \textbf{89.33\%$\pm$0.00\%} \\
3& New-Thyroid& 86.05\%$\pm$0.00\%& 86.05\%$\pm$0.00\%& 79.49\%$\pm$0.94       &\textbf{94.42\%$\pm$0.00\%} & 82.79\%$\pm$0.00\% & \textbf{97.21\%$\pm$0.00\%}\\
4& Dermathology &26.17\%$\pm$0.28\%& 61.67\%$\pm$0.26\%&  50.38\%$\pm$0.35     & \textbf{84.43\%$\pm$0.00\%} & 50.38\%$\pm$0.35\% & \textbf{84.70\%$\pm$0.00\%}\\
5& Lung-Cancer & 54.83\%$\pm$2.29\%&55.52\%$\pm$1.89\% & \textbf{74.48\%$\pm$3.33}      &{71.43\%$\pm$0.00\%} & \textbf{75.37\%$\pm$3.82\%} & \textbf{77.19\%$\pm$0.00\%} \\
6& COIL20(64x64) & 65.42\%$\pm$2.49\%& 69.15\%$\pm$2.38\%&  \textbf{79.56\%$\pm$2.72}      &\textbf{82.38\%$\pm$3.66\%} &  79.56\%$\pm$2.72\% &  \textbf{83.85\%$\pm$0.47\%}\\
7& COIL100 &49.75\%$\pm$1.31\%& 43.42\%$\pm$1.21\%&  {61.83\%$\pm$2.27}     & \textbf{68.93\%$\pm$1.98\%} &  61.83\%$\pm$2.27\% & \textbf{70.26\%$\pm$1.94\%}\\
8& MNIST(small) & 52.64\%$\pm$0.14\%& 34.49\%$\pm$0.10\%&  53.30\%$\pm$0.01      &\textbf{57.30\%$\pm$0.01\%} & 56.55\%$\pm$0.20\% & \textbf{62.28\%$\pm$0.00\%}\\
9& USPS & 43.70\%$\pm$2.84\%& 47.63\%$\pm$0.05\%&  \textbf{66.69\%$\pm$0.09}      &{60.43\%$\pm$0.90\%} & \textbf{66.69\%$\pm$0.09\%} & {61.93\%$\pm$1.01\%}\\
10& UMIST & 43.20\%$\pm$1.66\%& 43.44\%$\pm$1.92\%&  \textbf{70.99\%$\pm$2.19}      &{67.22\%$\pm$3.21\%} & \textbf{70.99\%$\pm$2.19\%}  & {67.22\%$\pm$3.21\%}\\
11& Extended-YaleB & 9.61\%$\pm$0.52\%& 10.59\%$\pm$0.49\%&  \textbf{34.74\%$\pm$0.90}      &{32.83\%$\pm$0.80\%} & \textbf{34.74\%$\pm$0.90\%} & {32.83\%$\pm$0.80\%} \\
12& ORL &54.37\%$\pm$2.41\%&  54.55\%$\pm$2.81\%& 57.33\%$\pm$2.37       &\textbf{63.93\%$\pm$2.67\%} & \textbf{68.85\%$\pm$2.48\%} & 66.20\%$\pm$1.46\%\\
\hline
summary&&&&&win:7; tied:1; lose:4&&win:6; tied:1; lose:5\\
\hline
\end{tabular}
\end{threeparttable}
}}
\end{table*}

   \begin{table*}[t]
\caption{\label{table:NMI_small}{NMI Comparison between the proposed method without parameter tuning and the well-tuned Gaussian RBF kernel.}}
\renewcommand{\arraystretch}{1.5}
\centerline{\scalebox{0.7}{
\begin{tabular}{llll}
\hline
& & Proposed$^{no\_tuning}$ & Spectral+Gaussian\_kernel{$^{optimal}$}\\
 \shline
1& Isolet1 & {78.93\%$\pm$0.44\%} & \textbf{79.66\%$\pm$0.58\%} \\
2& Wine& \textbf{63.94\%$\pm$0.00\%}& 43.02\%$\pm$0.00\% \\
3& New-Thyroid& \textbf{74.08\%$\pm$0.00\%}& 43.20\%$\pm$0.00\% \\
4& Dermathology & \textbf{88.61\%$\pm$0.00\%}& 51.04\%$\pm$0.15\%\\
5& Lung-Cancer & \textbf{55.70\%$\pm$0.00\%}& \textbf{55.58\%$\pm$4.60\%}\\
6& COIL20(64x64) & \textbf{93.54\%$\pm$1.26\%}& \textbf{93.29\%$\pm$0.81\%} \\
7& COIL100 & \textbf{90.45\%$\pm$0.25\%}& 89.28\%$\pm$0.59\% \\
8& MNIST(small) & \textbf{66.30\%$\pm$0.01\%}& 62.06\%$\pm$0.04\% \\
9& USPS & \textbf{66.75\%$\pm$0.84\%}& 66.05\%$\pm$0.18\%  \\
10& UMIST & \textbf{85.88\%$\pm$1.65\%}& \textbf{84.66\%$\pm$2.38\%}\\
11& Extended-YaleB & {44.30\%$\pm$0.72\%}& \textbf{47.11\%$\pm$0.58\%} \\
12& ORL &\textbf{82.72\%$\pm$0.97\%}& \textbf{83.62\%$\pm$1.13\%}\\
\hline
summary&&win:6; tied:4; lose:2&\\
\hline
\end{tabular}
}}
\end{table*}

   \begin{table*}[t]
\caption{\label{table:ACC_small}{ACC Comparison between the proposed method without parameter tuning and the well-tuned Gaussian RBF kernel.}}
\renewcommand{\arraystretch}{1.5}
\centerline{\scalebox{0.7}{
\begin{threeparttable}
\begin{tabular}{llll}
\hline
& & Proposed$^{no\_tuning}$ & Spectral+Gaussian\_kernel{$^{optimal}$}\\
 \shline
1& Isolet1    &{62.22\%$\pm$1.05\%} & \textbf{72.29\%$\pm$1.92\%}  \\
2& Wine    &\textbf{87.08\%$\pm$0.00\%} & 70.79\%$\pm$0.00\%  \\
3& New-Thyroid      &\textbf{94.42\%$\pm$0.00\%} & 82.79\%$\pm$0.00\%\\
4& Dermathology      & \textbf{84.43\%$\pm$0.00\%} & 50.38\%$\pm$0.35\%\\
5& Lung-Cancer      &{71.43\%$\pm$0.00\%} & \textbf{75.37\%$\pm$3.82\%} \\
6& COIL20(64x64)       &\textbf{82.38\%$\pm$3.66\%} &  \textbf{79.56\%$\pm$2.72\%} \\
7& COIL100     & \textbf{68.93\%$\pm$1.98\%} &  61.83\%$\pm$2.27\% \\
8& MNIST(small)       &\textbf{57.30\%$\pm$0.01\%} & 56.55\%$\pm$0.20\% \\
9& USPS      &{60.43\%$\pm$0.90\%} & \textbf{66.69\%$\pm$0.09\%} \\
10& UMIST       &{67.22\%$\pm$3.21\%} & \textbf{70.99\%$\pm$2.19\%} \\
11& Extended-YaleB      &{32.83\%$\pm$0.80\%} & \textbf{34.74\%$\pm$0.90\%} \\
12& ORL       &{63.93\%$\pm$2.67\%} & \textbf{68.85\%$\pm$2.48\%} \\
\hline
summary&&win:5; tied:1; lose:6&\\
\hline
\end{tabular}
\end{threeparttable}
}}
\end{table*}

\section{Conclusions}
In this paper, we have proposed a kernel learning method by a nonparametric density estimator. We have applied the kernel learning method to spectra clustering. Initial experimental results show that the proposed method outperforms the Gaussian RBF kernel, no matter whether the free parameters are fixed or not. Results further show that the proposed kernel learning method is insensitive to its free parameters.

\ifCLASSOPTIONcaptionsoff
  \newpage
\fi


\bibliographystyle{IEEEtran}

\bibliography{zxlrefs2}

\end{document}